\documentclass[10pt,twocolumn,letterpaper]{article}

\usepackage{cvpr}
\usepackage{times}
\usepackage{epsfig}
\usepackage{graphicx}
\usepackage{amsmath}
\usepackage{amssymb}
\usepackage{float}
\usepackage{subcaption}
\usepackage[numbers,sort]{natbib} %
\usepackage{booktabs}
\usepackage[pagebackref=true,breaklinks=true,letterpaper=true,colorlinks,bookmarks=false]{hyperref}

\hypersetup{
	plainpages=false,
	colorlinks=true,              %
	linkcolor=blue,               %
	anchorcolor=blue,             %
	citecolor=blue,               %
	filecolor=blue,               %
	urlcolor=blue,                %
	pdfview=FitH,                 %
	pdfstartview=FitH,            %
	pdfpagelayout=SinglePage      %
}

\newcommand\fig{Fig.}

\cvprfinalcopy %

\def\cvprPaperID{4057} %

\ifcvprfinal\fi
\begin{document}

\title{The Visual Centrifuge: Model-Free Layered Video Representations}

\author{Jean-Baptiste Alayrac\textsuperscript{1}\thanks{Equal contribution.}\\
{\tt\small jalayrac@google.com}
\and
Jo\~ao Carreira\textsuperscript{1}\footnotemark[1]\\
{\tt\small joaoluis@google.com}
\and
Andrew Zisserman\textsuperscript{1,2}\\
{\tt\small zisserman@google.com}
\and 
\textsuperscript{1}DeepMind
\quad
\textsuperscript{2}VGG, Dept.\  of Engineering Science, University of Oxford
}

\maketitle
\begin{abstract}
True video understanding requires making sense of non-lambertian scenes where the color of light arriving at the camera sensor encodes information about not just the last object it collided with, but about multiple mediums -- colored windows, dirty mirrors, smoke or rain. Layered video representations have the potential of accurately modelling realistic scenes but have so far required stringent assumptions on motion, lighting and shape. Here we propose a learning-based approach for multi-layered video representation: we introduce novel uncertainty-capturing 3D convolutional architectures and train them to separate blended videos. We show that these models then generalize to single videos, where they exhibit interesting abilities: color constancy, factoring out shadows and separating reflections. We present quantitative and qualitative results on real world videos.
\end{abstract}

\section{Introduction}
Vision could be easy and require little more than mathematical
modelling: the brightness constancy constraint for optical flow,
sobel-filtering and perspective equations for 3D object recognition,
or lambertian reflectance for shape-from-shading. However, the messiness of the real world has long proved the
assumptions made by these models inadequate: even simple natural
scenes are riddled with shadows, reflections and colored lights that
bounce off surfaces of different materials and mix in complex
ways. Robust systems for scene understanding that can be safely
deployed in the wild (e.g.\ robots, self-driving cars) will probably
require not just tolerance to these factors, as  current deep
learning based systems have; they will require factoring out these
variables in their visual representations, such that they do not get
rattled by giant (reflected) trees growing out of potholes in the
road, or even their own shadow or reflection.

A natural framework for handling these factors is to model them as
layers that compose into an overall video. Layered models trace back
to the foundations of computer vision \cite{Wang1994RepresentingMI} but assumed particular models of motion \cite{jojic2001learning},
scene shapes or illumination. Layered models are also
often tailored to particular goals -- such as shadow or specularity removal,
 or reflection separation \cite{Szeliski2000} and
rarely accomodate for non-rigidity other than for very specialized
domains (e.g.\ faces \cite{liu2017better}).
\begin{figure}[t]
	\begin{center}
		\includegraphics[width=\linewidth]{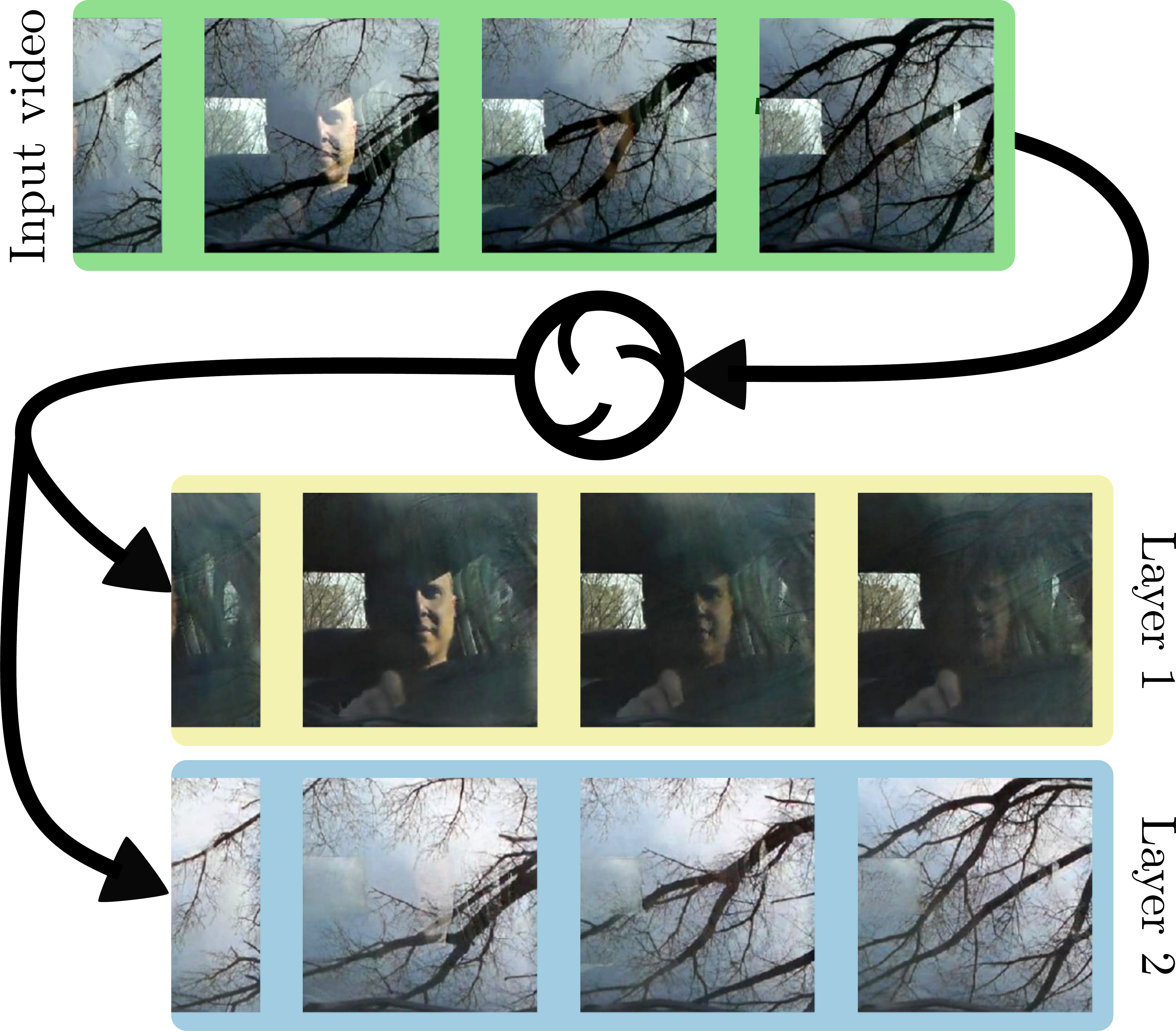}
	\end{center}
	\caption[]{\label{fig:street_scene} \small Top, input video showing someone driving a car through the country side, with trees reflecting in its windscreen. Bottom, two videos output by our \textit{visual centrifuge}\footnotemark. In this paper we learn models that can, in the spirit of a centrifuge, separate a single video into multiple layers, \eg to consider the interior of the car or the shapes of the reflected trees in isolation. We do so using few assumptions, by simply training models to separate multiple blended videos -- a task for which training data can be obtained in abundance.}
\end{figure}
\footnotetext{See \url{https://youtu.be/u8QwiSa6L0Q} for a video version of the figure.}

In this paper we aim to learn a video representation that teases apart
a video into layers in a more general data-driven way that does away
with explicit assumptions about shape, motion or illumination. Our
approach is to train a neural network that, in the spirit of a
\textit{visual centrifuge},
separates pairs of videos that we first blend together using uniformly
weighted averaging. Related ideas have been pursued in the audio
domain~\cite{Yu2017,Afouras2018,Ephrat2018}, where signals are waves that really combine
additively by superposition. In the visual domain this approximation is accurate when
dealing with some reflections but not necessarily in other cases of
interest such as shadows or extremely specular surfaces such as
mirrors. However, we hope that by mixing a sufficiently large and
diverse set of videos these cases will also be sporadically
synthesized and the model can learn to separate them (e.g.\ a shadow
from one video will darken the blended video, and needs to be factored
out to reconstruct the second video).

How is it possible for the network to separate the mixture into its constituent videos? There are two
principal cues that can be used: the different motion fields of the two videos, and the semantic content, e.g.\ picking out a car  in one video and a cow in another. There are also more subtle cues such as one `layer' may be more blurred or have different colors. 
We show that our model, after being trained on blended pairs of
videos from Kinetics-600~\cite{kay_arxiv_2017,Carreira-Kinetics600-2018}, a large video dataset with around
400k 10-second clips of human actions, can indeed spontaneously separate natural reflections
and shadows as well as remove color filters from new individual
(non-blended) videos as shown in \fig~\ref{fig:street_scene}. 

While our model is not necessarily more accurate
than existing ones on individual niche tasks in constrained settings,
although it has comparable performance, it can also succeed on a variety of layer separation tasks in totally
unconstrained settings where previous models fail (e.g.\ with people moving around and shaky
cameras).
\\

\vspace{1mm}
\noindent\textbf{Contributions.}
Our contributions are threefold; \textbf{(i)} we propose novel architectures for multi-layered video modelling,
\textbf{(ii)} we show that these models can be learned without supervision, by just separating synthetically blended videos and,
\textbf{(iii)} we observe that these models exhibit color constancy abilities and can factor out shadows and reflections on real world videos.
\\

\section{Related work}
\label{sec:RW}

\noindent\textbf{Image layer composition.}
Many different layer composition types have been developed that model the image generation process. 
Intrinsic image approaches~\cite{Barron2015,Fan2018,Finlayson2014,Sinha1993,Tappen2002,Weiss2001} aim to factorize illumination, surface reflectance and shape. Deconvolution algorithms, such as blind deblurring, model the image as a superposition of multiple copies of an original (unblurred) 
image~\cite{Cho07,Fergus06,Shan08,Yuan07,Whyte12,Jin2018LearningTE}.
A related problem is the one of color constancy~\cite{Barron2015a, Barron2017}, where the goal is to infer the color of the light illuminating a scene in order to remove it.
\noindent\textbf{Reflection removal.}
Reflections in natural images are a particular case of layer composition, where two or more layers are mixed together through simple addition in order to form the final image.
Most successful classical methods for removing reflections assume access to a sequence of images where the reflection and the background layer have different motions~\cite{Beery2008, Guo2014, KunGai2012, Szeliski2000, Xue2015, Nandoriya2017}.
By recovering the two dominant motions, these methods can recover the original layers through temporal filtering.
The work by Xue et al.~\cite{Xue2015} notably proposes an optimization procedure which alternates between estimating dense optical flow fields encoding the motions of the reflection and the background layers and recovering the layers themselves, which leads to impressive results on images containing natural reflections.
However, all these methods rely on the assumption that the two layers have distinctive and almost constant motions~\cite{Xue2015,Guo2014} and cannot handle cases where multiple objects are moving with independent motions inside the layers. 

Recently, Fan et al.~\cite{Fan2017} proposed a deep learning architecture to suppress reflections given a single image only.
The advantage of this and related approaches \cite{zhang2016colorful,Chi2018,Zhang2018SingleIR} is that they are very flexible -- given appropriate data they can in principle operate in unconstrained settings.

\noindent\textbf{Video layer decomposition.}
All previously mentioned approaches are designed to output results for one image. 
We focus instead on recovering layers composing a whole video~\cite{kumar2008learning,jojic2001learning}. 
As observed in~\cite{Nandoriya2017}, simple extensions of the previous techniques to videos, such as applying the methods in a frame by frame fashion followed by temporal filtering is not satisfactory as it leads to strong temporal flickering, incomplete recovery of the layers and often blurs the objects present in the video.
To alleviate these issues,~\cite{Nandoriya2017} propose an extension of the work of~\cite{Xue2015} but where they adapt both the initialization strategy and the optimization objective in order to take into account the temporal dimension.
The proposed approach strongly alleviates the temporal flickering issue.
However, the method still relies on strong assumptions concerning the relative motions of the two layers and might notably suffer from objects moving fast inside one of the layers.
Differently from~\cite{Nandoriya2017}, we want to rely on semantic cues whenever motion cues are not sufficient.

\begin{figure*}[t!]
	\centering
	\begin{subfigure}[t]{.27\linewidth}
		\centering
		\includegraphics[height=3.3cm]{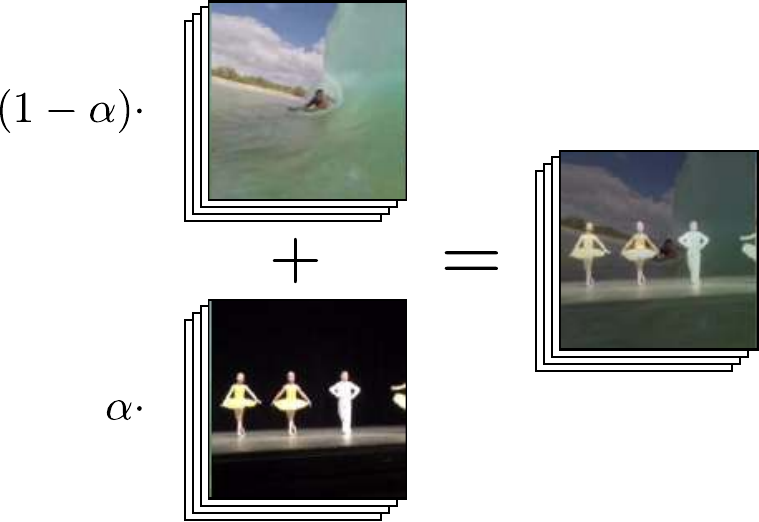} %
		\caption{\small \textbf{Video generation} (\ref{subsec:generation}) \label{fig:generation}}
	\end{subfigure}
	\hfill
	\begin{subfigure}[t]{.45\linewidth}
		\centering
		\includegraphics[height=3.3cm]{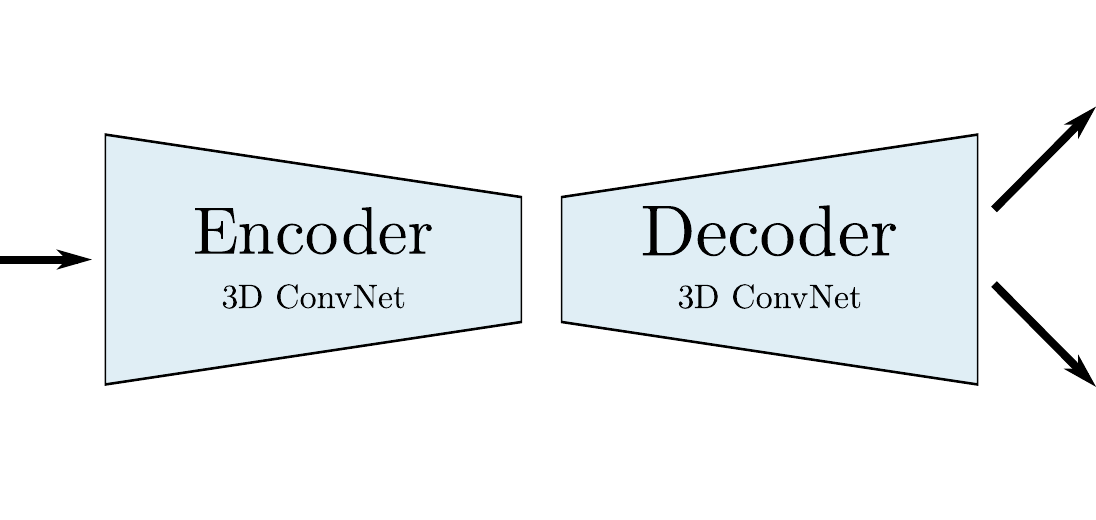} %
		\caption{\small \textbf{Model architecture} (\ref{subsec:architecture}) \label{fig:architectures}}
	\end{subfigure}
	\hfill
	\begin{subfigure}[t]{.27\linewidth}
		\centering
		\includegraphics[height=3.3cm]{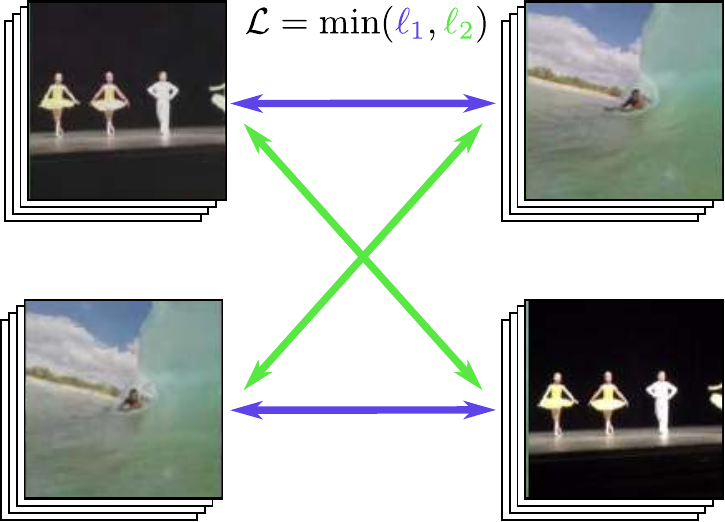} %
		\caption{\small \textbf{Permutation invariant loss} (\ref{subsec:pil}) \label{fig:losses}}
	\end{subfigure}
\vspace*{-0.3cm}
	\caption{\label{fig:model}
		Illustration of the general idea, described in full detail in section~\ref{sec:model}. Two videos are blended together into a single video and this video is passed through a neural network which is trained to separate it back into the two original videos. The hope is that the underlying learned representation captures the concept of natural video layers and that it will then generalize when processing standard videos. Real separation results shown.
	}	
\end{figure*}

\noindent\textbf{Permutation invariant losses.}
We want to recover the layers composing the image in a blind manner, \ie without making assumptions about the different layers nor giving external cues that could indicate which layer we want to reconstruct.
This is challenging as it relates to the label permutation problem~\cite{Yu2017}.
One solution to this problem proposed in the audio domain is to make use of permutation invariant losses~\cite{Yu2017}.
Here we employ a similar strategy by adapting this principle to video reconstruction.
This is also related to the problem of uncertainty
inherent to the fact that multiple solutions are possible for the
layer decomposition, a situation that can be handled by
designing a network to  generate multiple hypotheses, with an appropriate loss to reward only one for
each training sample~\cite{Rupprecht2017,firman2018diversenet,li2018interactive}.
In this work we propose to use permutation
invariant losses in the context of emiting multiple hypotheses for the
layers.

\noindent\textbf{Audio separation.}
Additive layer composition is particularly suitable for modeling how
different audio sources are assembled to form a sound.  That is why
our work also relates to the audio separation domain -- in particular to `blind
source separation'. However, much of the literature on blind audio separation, such as for
the well known  `Cocktail Party problem',  requires multiple audio channels (microphones) 
as input~\cite{Comon10}, which is not the situation we consider in this work. 
Though deep learning has brought fresh interest in the single audio channel  case, 
e.g.~\cite{Erdogan2015PhasesensitiveAR,Wang2018EndtoEndSS}.
Recent work
has  revisited the cocktail party
problem while also using visual cues~\cite{Afouras2018, Ephrat2018,Gao2018LearningTS}.

\noindent\textbf{Layers beyond graphics.} 
Others have also investigated the use of image layers composition for other purposes than computers graphic applications.
For example, recent work explores additive layer composition as a data augmentation technique for image level classification~\cite{Inoue2018, Tokozume2018, Zhang2017}.
Interestingly, \cite{Zhang2017} shows that simply mixing images and labels in an additive fashion improves generalization and robustness to adversarial examples as well as stabilizes training of generative models.
Such techniques have not yet been extended to the video domain as we do in this work.

\section{Deep layer separation by synthetic training}
\label{sec:model}
In this section we describe our model which is trained end-to-end to reconstruct layers composing an input video.
We generate the training data synthetically using a simple additive layer composition as explained in section~\ref{subsec:generation}.
In section~\ref{subsec:architecture}, we describe the  model architecture  to tackle our problem.
Finally, we motivate our choice of loss in section~\ref{subsec:pil}.
\fig~\ref{fig:model} summarizes our approach.

\subsection{Video generation process}
\label{subsec:generation}

Real videos with ground truth layer decomposition are hard to get at scale.  
To be able to train a neural network for this task, we instead generate artificial videos for which we have easily access to ground truth.
In practice, we average two videos with various coefficients, a simple strategy already proposed in~\cite{Szeliski2000} to evaluate image decomposition models.
More formally, given two videos $V^{1}, V^{2}\in\mathbb{R}^{T\times H \times W \times 3}$,  where $T$ is the total number of frames, $H$ and $W$ the frame's height and width and $3$ corresponds to the standard RGB channels, we generate a training video $V$ as follows:
\begin{equation}
\label{eq:generation}
	V = (1-\alpha) \cdot V^{1} + \alpha \cdot V^{2},
\end{equation}
where $\alpha\in\left[0,1\right]$ is a variable mixing parameter.
This process is illustrated in \fig~\ref{fig:generation}.

Despite this apparent simple data generation scheme, we show in
section~\ref{sec:applications} that this is sufficient to train a  model
that can generalize to real videos with layer composition including
shadows and reflections.

\subsection{Model architecture}
\label{subsec:architecture}
We use an encoder-decoder type architecture that, given an input mixed video, outputs two or more videos aiming at recovering the original layers composing the input (see \fig~\ref{fig:architectures}).
We denote by $V$ the input video and by $O$ the $n$ outputs of the network, where $O^i$ corresponds to the $i$-th outputed video. 
Below, we give details about our particular design choices.

\noindent\textbf{3D ConvNet.}
As demonstrated by previous work~\cite{Xue2015}, motion is a major cue to reconstruct the composing layers.
For this reason, we leverage a 3D ConvNet architecture able to capture both appearance and motion patterns 
at multiple temporal scales to succeed at the task.
For the encoder, we use the I3D architecture~\cite{carreira17quovadis} which has proven to be effective for video classification.
For the decoder, we propose a simple architecture which consists of a succession of 3D Transposed Convolutions~\cite{Dumoulin2016} that we detail in Appendix~\ref{app:architecture}.

\noindent\textbf{U-Net.}
To improve the quality of reconstruction, we follow  the U-Net architecture~\cite{Ronneberger2015} that has proved its worth in many dense reconstruction tasks, e.g.~\cite{Isola2017ImagetoImageTW}, 
and add skip connections between the encoder and the decoder (see Appendix~\ref{app:architecture} for details).

\noindent\textbf{Output layers.} 
Although our synthetic video are composed by mixing only two videos,
we found it helpful to allow our models to produce more than two outputs. This
is to alleviate the problem of uncertainty~\cite{Rupprecht2017}
inherent to our task, \ie multiple solutions for the layers are often
possible and satisfactory to reconstruct the input. To output $n$ videos, we simply
increase the number of channels at the output of the network; given a
video $V\in\mathbb{R}^{T\times H \times W \times 3}$, the network is
designed to output $O\in\mathbb{R}^{T\times H\times W\times 3n}$.  This
means that the separation of the outputs only happens at the end of the
network, which makes it possible for it to perform quality verification along the way (e.g. check that the  outputs sum correctly to the input).  Although introducing multiple alternative outputs may lower applicability in some cases, simple
strategies can be adopted to automatically choose two outputs out of
$n$ at test time, such as selecting the two most dissimilar video layers (which we do 
by selecting the most distant outputs in pixel space).

\noindent\textbf{Predictor-Corrector.}
We also give the model the possibility to further correct its initial predictions by stacking a second encoder-decoder network after the first one.
This is inspired by the success of iterative computation architectures~\cite{carreira2016human,Newell2016} used in the context of human pose estimation.
Given an initial input mixed video $V\in\mathbb{R}^{T\times H \times W \times 3}$ and $n$ target output layers, the first network, the \emph{predictor}, outputs an initial guess for the reconstruction $\tilde{O}\in\mathbb{R}^{T\times H \times W \times 3n}$.
The second network, the \emph{corrector}, takes $\tilde{O}$ as input and outputs $\Delta\in\mathbb{R}^{T\times H \times W \times 3n}$ such that the final output of the network is defined as
$
O = \tilde{O}+\Delta.
$
Because the role of these two networks are different, they do not share weights.
We train the two networks end-to-end from scratch without any specific two-stage training procedure.

\subsection{Permutation invariant loss}
\label{subsec:pil}
One challenge of our approach lies in the fact that we do not have any a priori information about of the order of the input video layers.
Therefore it is hard to enforce the network to output a given layer at a specific position. This challenge is usually refered as the permutation label problem~\cite{Yu2017}.

To overcome this problem, we define a training loss which is permutation invariant (see~\fig~\ref{fig:losses}).
More formally, given the two original ground truth videos $\{V^{1},V^{2}\}$ and the outputs of our network $O$ defined previously, we set up the training loss as:
\begin{equation}
\label{eq:training_loss}
\mathcal{L}\left(\{V^{1}, V^{2}\}, O\right)=\min_{(i,j) | i\neq j} \ell(V^{1}, O^i)+\ell(V^{2}, O^j),
\end{equation}
where $\ell$ is a reconstruction loss for videos.

Following previous work~\cite{Mathieu2016}, we define $\ell$ for two videos $U$ and $V$ as follows:
\begin{equation}
 \label{eq:recons_loss}
 \ell(U, V) = \frac{1}{2T} \left(\sum_t \Vert U_t-V_t \Vert_1 + \Vert \nabla(U_t)-\nabla(V_t) \Vert_1\right),
\end{equation}
where $\Vert \cdot \Vert_1$ is the $L_1$ norm and $\nabla(\cdot)$ is the spatial gradient operator.
We noticed that adding the gradient loss was useful to set more emphasis on edges which were usually harder to capture when compared to constant areas.

\section{Experiments}
\label{sec:experiments}

We trained models on the task of unmixing averaged pairs of videos then tested these models on individual videos from the web and in the wild. The models were trained on pairs of videos from the Kinetics-600 dataset~\cite{Carreira-Kinetics600-2018} training set, which has approximately 400k 10s long videos (250 frames). We evaluated generalization on the Kinetics-600 validation set, which has 30k videos. We used standard augmentation procedures: random left-right flipping and random spatiotemporal cropping, where the shortest side of the video was first resized to 1.15x the desired crop size. Most of the experiments used 32-frame clips with 112x112 resolution for fast iteration. We also trained the full proposed architecture on 64-frame clips with 224x224 resolution -- we report results with this model in the applications section~\ref{sec:applications}.
We tried sampling the blending parameter $\alpha$ of Eq.~\eqref{eq:generation} in $\left[0.25,0.75\right]$ without observing a strong influence on the results when compared to fixed sampling scheme.
Therefore, we simply use $\alpha=0.5$.

\subsection{Architecture Evaluation} 

Here we compare the performance of multiple architectural variations
on the learning task of separating averaged videos. 
We first evaluate using the reconstruction loss, and then use a downstream task -- that of human action recognition.
All architectures
share the same basic predictor module. All models were trained using
SGD with momentum, with the same hyperparameters: learning rate 0.1,
momentum 0.99, no weight decay and batch size of 10 clips. The
learning rate is lowered to 0.05 at 100k iterations, 0.025 at 150k and
0.01 at 200k. The models are trained for a total of 240k
iterations. At test time moving averages are used in batch
normalization layers.

The first observation was that even the simplest model works: using the permutation-invariant loss, the blended videos separate into the original ones. The loss of the basic predictor model with two output video layers, is provided in table~\ref{table:baselines} and can be contrasted with two baselines: 1) outputing twice the blended video, 2) outputting two different layers, but using the predictor with random weights (no training). The loss of the trained model is significantly lower, although the layers are still somewhat noisy. Our more advanced models are more accurate.

\begin{table}[ht]
	\centering
\begin{tabular}{@{}cc@{}}
	\toprule
	Model                   & Validation loss \\ \midrule
	Identity                & 0.361           \\
	Predictor (no training) & 0.561           \\
	Predictor (trained)     & 0.187           \\ \bottomrule
\end{tabular}
\caption{\label{table:baselines} \small Validation losses obtained by the basic predictor -- an encoder-decoder model producing two output layers. \textit{Identity} is a baseline where the two output video layers are just copies of the input blended video. The second baseline is the predictor without any training, using the initial random weights. } 
\end{table}

\begin{figure}[t]
	\begin{center}
		\includegraphics[width=\linewidth]{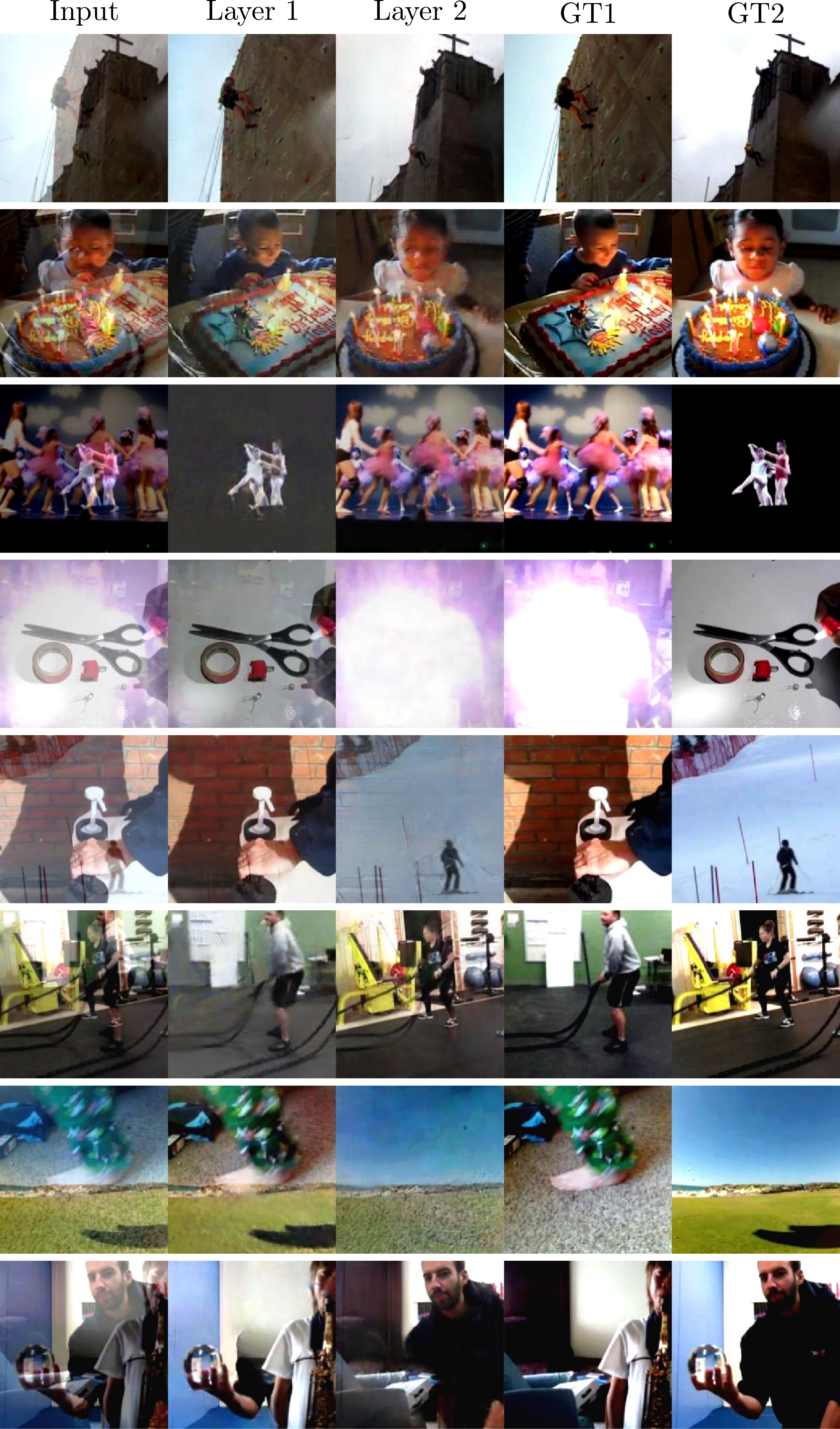}
	\end{center}
	\vspace*{-0.5cm}
	\caption{\label{fig:standard_results} \small Example outputs of the model on blended Kinetics validation clips. Due to lack of space we show a single frame per clip. Original unblended videos are shown on the rightmost columns. Overall the network is able to unmix videos with a good accuracy even when confronted with hard examples, \eg, videos from the same class. The first five rows show successful separations. The last three show rare cases where the network cuts and pastes in a coherent manner some objects between videos.}
		\vspace*{-0.2cm}
\end{figure}

We also found that predicting more than 2 layers for each video results in substantially better unmixing -- 
we observed that often the outputs formed two clear clusters of video layers, and that 
two of the  layers among the predicted set are considerably more accurate than those obtained when predicting just $2$ overall. 
These results can be found in table~\ref{table:analysis}, second column. %
We think that producing additional layers is mainly helping the
training process by allowing the model to hedge against factors like
differences in brightness, which may be impossible to invert, and to
focus on separating the content (objects, etc.).

Table~\ref{table:analysis} also shows the benefits of the predictor-corrector architecture, using a single correction module, especially when predicting multiple (more than 2) video layers. It is also likely that additional correction steps would improve performance further -- we plan to verify this in future work. %
The results in the rest of the paper used the predictor-corrector architecture with $4$ output video layers.

\begin{table}[ht]
\centering 
\begin{tabular}{@{}ccc@{}} 
\toprule
\# output video layers & Predictor & Predictor-Corrector \\ \midrule
2  & 0.187 & 0.172\\ 
4 &  0.159 & 0.133\\
8 & 0.151 & - \\
12 & 0.150 & - \\ \bottomrule
\end{tabular}
\caption{\label{table:analysis} \small  Validation loss when producing various number of output video layers, for a simple predictor model and for the predictor-corrector model. Larger sets of layers tend to contain higher-quality reconstructions of the original videos, but this starts to saturate at around 4 -- there is little improvement when increasing to 8 or 12 for the predictor model and we did not experiment with such large sets of output layers on the more memory-demanding predictor-corrector model. Finally, the predictor-corrector model outperforms the predictor by a significant margin, especially when computing $4$ output video layers.} 
\end{table}

\noindent
\textbf{Additional  loss functions.}  We mention here two loss functions that we experimented with,  but that
ultimately brought no benefit and are not used. First, it might be  expected that it is important to
enforce that the output layers should recompose into the original mixed video as a consistency check. This can
be achieved by adding a loss function to the objective:
 \begin{equation}
 \ell(V, (1-\alpha)\cdot O^i + \alpha\cdot O^j),
 \end{equation}
where $i$ and $j$ are respectively the indexes of the layers matched to $V_1$ and $V_2$ 
according to equation~(\ref{eq:training_loss}).
However,  we did not observe an  outright improvement -- possibly because for real sequences (see below) the 
strict addition is only a weak model for layer formation.  
We also considered 
enforcing diversity in the outputs through an explicit loss term, 
$-\ell(O^i,O^j)$.
This also did not bring immediate improvement 
(and without reconstruction constraints and proper tuning  was generating absurdly diverse outputs).
Note also that in general the outputs are diverse when measured with simple diversity losses,
despite some small cross-talk, so more efforts might be needed to design a more appropriate diversity loss.

\noindent
\textbf{Evaluating on a downstream task.}
We evaluated the quality of the separated videos for the task of human action recognition.
To this end,  we tested I3D (that has been trained on the standard Kinetics training set):  (a) directly on mixed pairs of videos, (b) on \textit{centrifugally}-unmixed pairs of videos, and (c) on the original clean pairs of videos on the validation set of the Kinetics dataset 
(using only 64-frames clips for simplicity, though better results can be obtained on the full 250-frame clips).
We used  a modified version of accuracy for evaluation -- as we have two different videos, 
we allow the methods to make two predictions.
We consider a score of $1$ if we recover the two ground truth labels, a score of $0.5$ if we recover only one of the two labels and a score of $0$ otherwise.
For method (a), we simply take its top 2 predictions.
For method (b) and (c), we take the top-1 predictions of the two branches.
In this setting, the centrifuge process improved accuracy from $22\%$ for (a)  to $44\%$ for (b). However, there is still a gap with the original setup (c) which achieves an accuracy of $60\%$. The gap is presumably due to persistent artifacts in the unmixed videos.

\subsection{Psychophysics of Layered Representations}

Having established the benefits of our proposed architecture, it is interesting to probe into it and see what it has learned, its strengths and weak points which we attempted to do by running a series of psychophysics-like experiments.

\begin{figure}[t]
\begin{center}
\includegraphics[width=\linewidth]{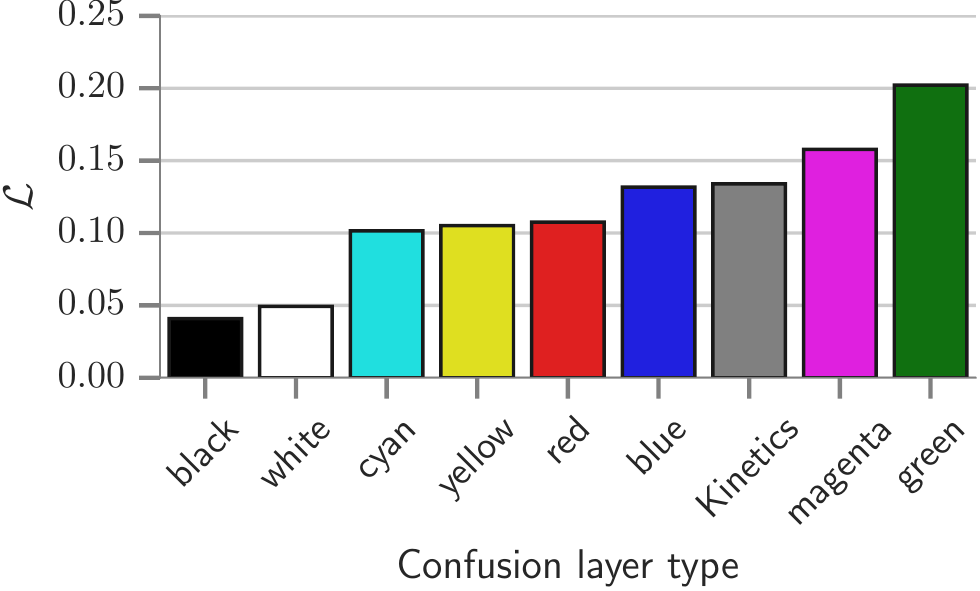}
\end{center}
\vspace*{-0.5cm}
\caption{\label{fig:color_const_barplot} \small Loss obtained by the predictor-corrector model when separating Kinetics videos from pure-colored video of different hues. The loss obtained when separating pairs of Kinetics-videos is shown for reference as the gray bar -- note that the while the model is accurate at separating pairs of Kinetics videos, for which it was explicitly trained, it is even better at separating most of these pure-colored videos, a task for which it was not trained for. Some colors, however, make the task quite hard -- magenta and green, perhaps due to less frequent in the natural videos from Kinetics.}
\end{figure}

\begin{figure}[t]
\begin{center}
\includegraphics[width=\linewidth]{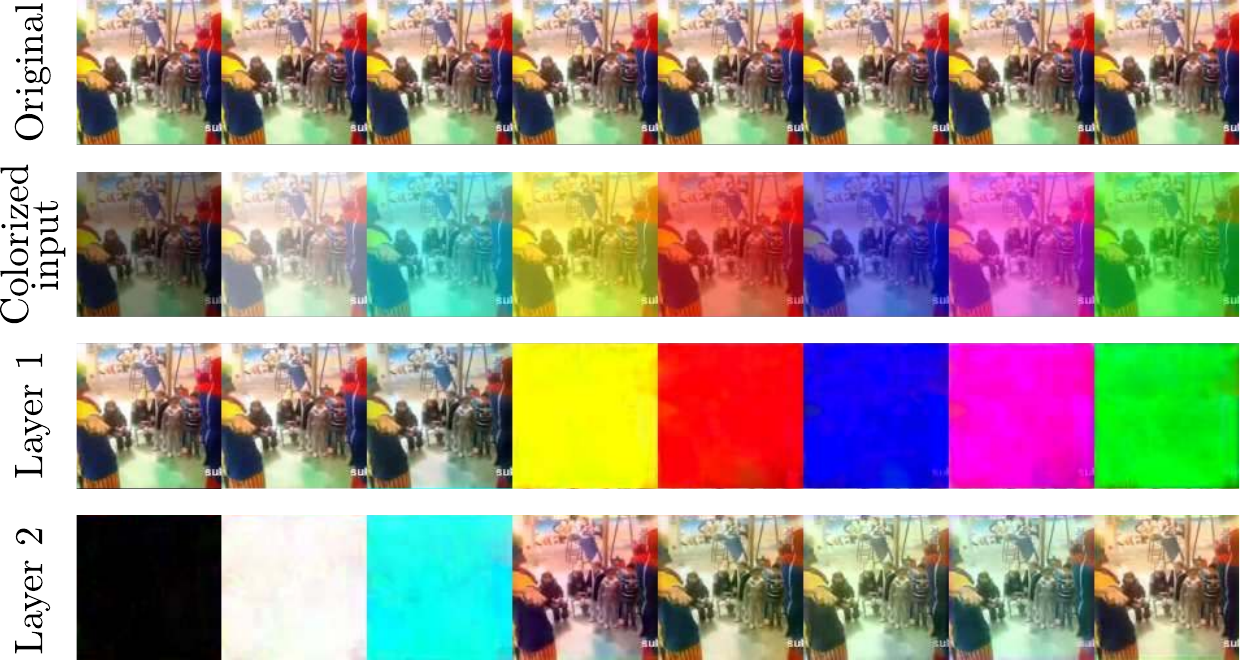}
\end{center}
\vspace*{-0.5cm}
	\caption{\label{fig:color_const} \small Top: frame from original video. 2nd row: same frame from same video after mixing with different colored videos. 3rd and 4th rows: $2$ video layer outputs from our predictor-corrector. Note that the reconstructions of the original video are quite similar and that the colored layers are also well reconstructed, despite a highly colorful scene (e.g.\  the clown's shirt has yellow sleeves).}
\vspace*{-0.5cm}
\end{figure}

\noindent \textbf{Color.} In human vision the colors of objects are perceived as the same across different lighting conditions -- independently of whether the sun is shining bright at mid-day, or nearing sunset and factoring out any cast shadows. We experimented with an extreme notion of color constancy and transformed Kinetics videos as if they had been captured by cameras with different pure-colored filters: black, white, green, red, yellow, blue, cyan and magenta, by averaging them with \textit{empty} videos having just those colors. We did not train on this data, instead we simply used the best model trained to separate pairs of Kinetics videos. We observed that the model generalized quite well to these videos and did accurately reconstruct the two layers in most cases -- one Kinetics video and one pure color video -- and the results are shown in \fig~\ref{fig:color_const_barplot}. It can be seen that the task is easier for black and white filters, which is natural since it corresponds roughly to just darkening or brightening a video. The hardest cases are magenta and green filters, perhaps because these colors are less common in our training data -- we leave this analysis for future work, the main point here being that the models generalize well to very different layer compositions. Results for an example frame are shown in \fig~\ref{fig:color_const}.

\noindent \textbf{Motion vs.\ Static Cues.} Motion plays a critical role in engineered solutions (in constrained settings) to problems such as reflection removal (e.g.\  \cite{Xue2015}). To understand how important motion is in our models, compared to static scene analysis, we trained a second predictor-corrector model with $4$ output layers, using the exact same experimental setting as before, but now training on videos without motion. We generated these \textit{frozen} videos by sampling a frame from a normal video and repeating it $32$ times to get each $32$-frame clip. We then evaluated the two models on both normal and frozen videos to see how they generalize. We also tried mixing pairs composed of one normal video and one frozen video. The $6$ different val. losses appear in table \ref{table:frozen_normal}. 

\begin{figure}[t]
	\begin{center}
		\includegraphics[width=\linewidth]{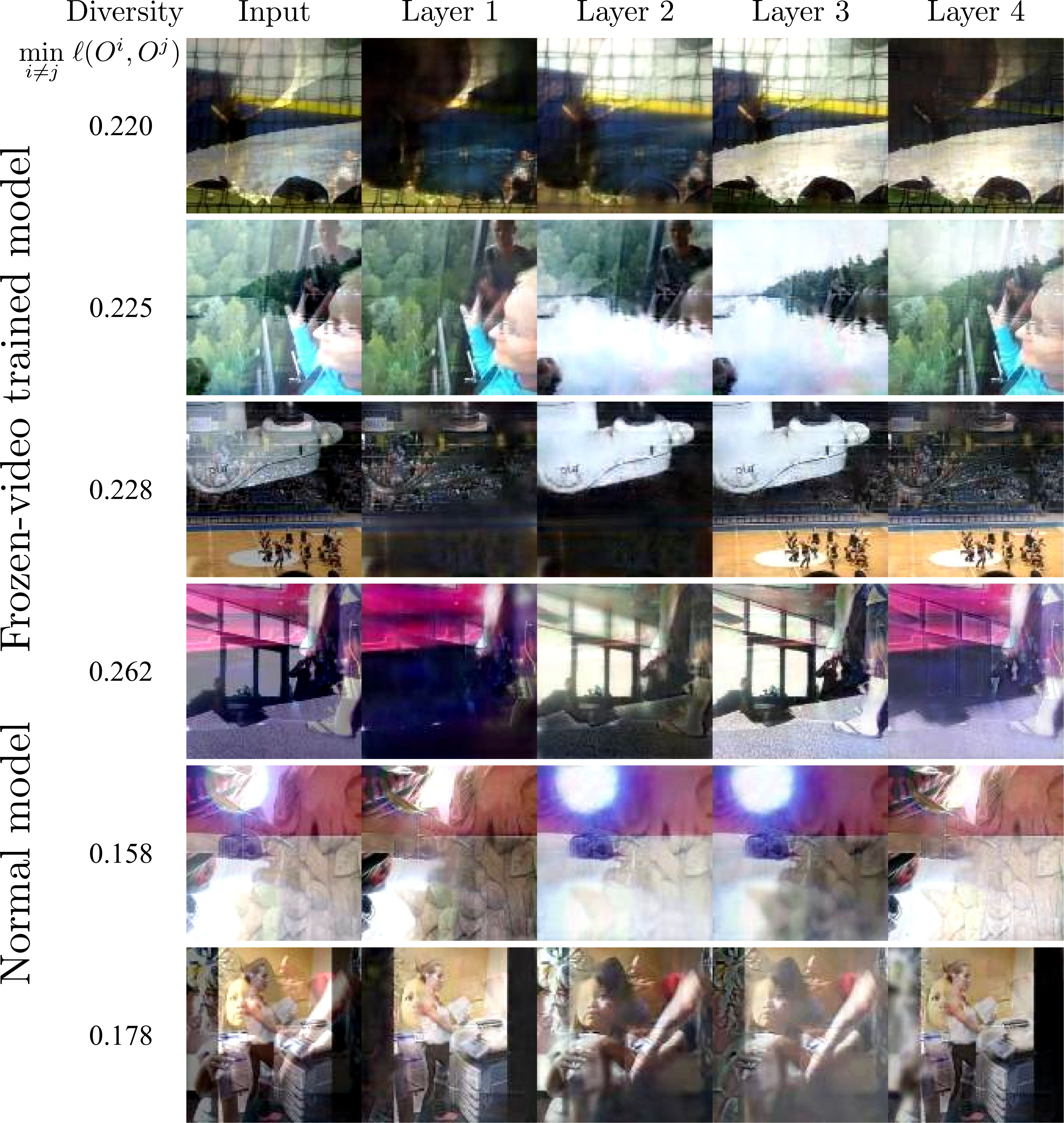}
	\end{center}
	\vspace*{-0.5cm}
	\caption{\label{fig:diversity} \small 
	Example videos where our models produce highly diverse sets of layers. 
	\textbf{Top 4 rows}: layers output by a model trained and tested on frozen videos;
	\textbf{Bottom 2 rows}: layers output by a model trained and tested on regular videos. In both cases we sort the videos by layer diversity (from least diverse on the top to most diverse on the bottom).
	We observe that the diversity in output video layers is much higher for the model on frozen videos -- motion is a strong cue to disambiguate between the layers.
	Note that we selected these blended videos automatically by blending many pairs and choosing here the ones that maximize the diversity metric, $\min_{i\neq j} \ell(O^i, O^j)$ (shown on the left), over 1K runs.	
	}
\end{figure}

We found that motion is an important cue in our system: it is harder to unmix frozen videos than motion ones. Also, the system trained on motion videos is worse on mixed frozen videos than the model trained on frozen videos. However, if just one of the videos is frozen then the motion-trained model excels and does better even than when both videos have motion -- perhaps during training the model receives some examples like. Finally, the model trained on frozen videos does poorly when processing inputs which contain motion -- this is natural, since it never seen those in training. Interestingly, we noticed also that the sampled layers tend to be significantly more diverse for frozen videos, reflecting the fact that they are more ambiguous.
To further support that point, we computed an average diversity metric, $\min_{i\neq j} \ell(O^i, O^j)$, over 1K runs.
For the frozen video model on frozen videos, we obtained an average diversity score of $0.079$ versus $0.045$ for our standard model on motion videos.
\fig~\ref{fig:diversity} shows outputs with maximum diversity score for both models.

\begin{table}[ht]
\centering 
\begin{tabular}{@{}cccc@{}} 
\toprule
Train/Test & 2 frozen & 2 normal & 1 frozen 1 normal\\ 
\midrule
2 frozen & 0.165 & 0.233 & 0.198 \\ 
2 normal & 0.205 & 0.133 & 0.127 \\
\bottomrule
\end{tabular}
\caption{\label{table:frozen_normal} Loss obtained when training/testing on pairs of frozen/normal videos, and testing on pairs of frozen/normal videos or when blending one frozen and one normal video. A frozen video is a video obtained by just repeating many times a single frame from a normal video, such that it does not have motion.} 
\end{table}

\begin{table}[ht]
	\centering 
	\small
	\begin{tabular}{@{}ccc@{}}
		\toprule
		Encoder endpoint & depth & Validation loss \\ \midrule
		Mixed\_3c        & 7     & 0.214           \\
		Mixed\_4f        & 17    & 0.181           \\
		Mixed\_5c        & 21    & 0.187           \\ \bottomrule
	\end{tabular}

\caption{\label{table:depth} \small Validation losses obtained when using three increasingly deeper subnetworks of the I3D encoder. The two deeper models achieve lower loss, indicating the value of high-capacity and wide receptive fields in space and time on this task.} 

\end{table}

\noindent \textbf{Low-level vs.\ high-level features.} Another interesting question is whether the models are relying on high-level semantic cues (e.g. people's shapes, scene consistency) or just on low-level ones (e.g. texture, edges, flow). We ran several experiments to try to shed light on this.

First, we revisited the basic predictor model and varied the depth of the encoder architecture, by taking features from three different layers of I3D: ``Mixed\_3c", ``Mixed\_4f"  and ``Mixed\_5c" (the default elsewhere in the paper). These correspond respectively to encoders with 7, 17 and 21 convolutional layers. The results in table~\ref{table:depth} show that the two deeper encoders perform considerably better than the shallower one, suggesting that higher-level, semantic features matter, but this may also be due to greater fitting capacity and/or larger spatio-temporal receptive fields being required.

\begin{figure*}[t!]
	\begin{center}
		\includegraphics[width=\textwidth]{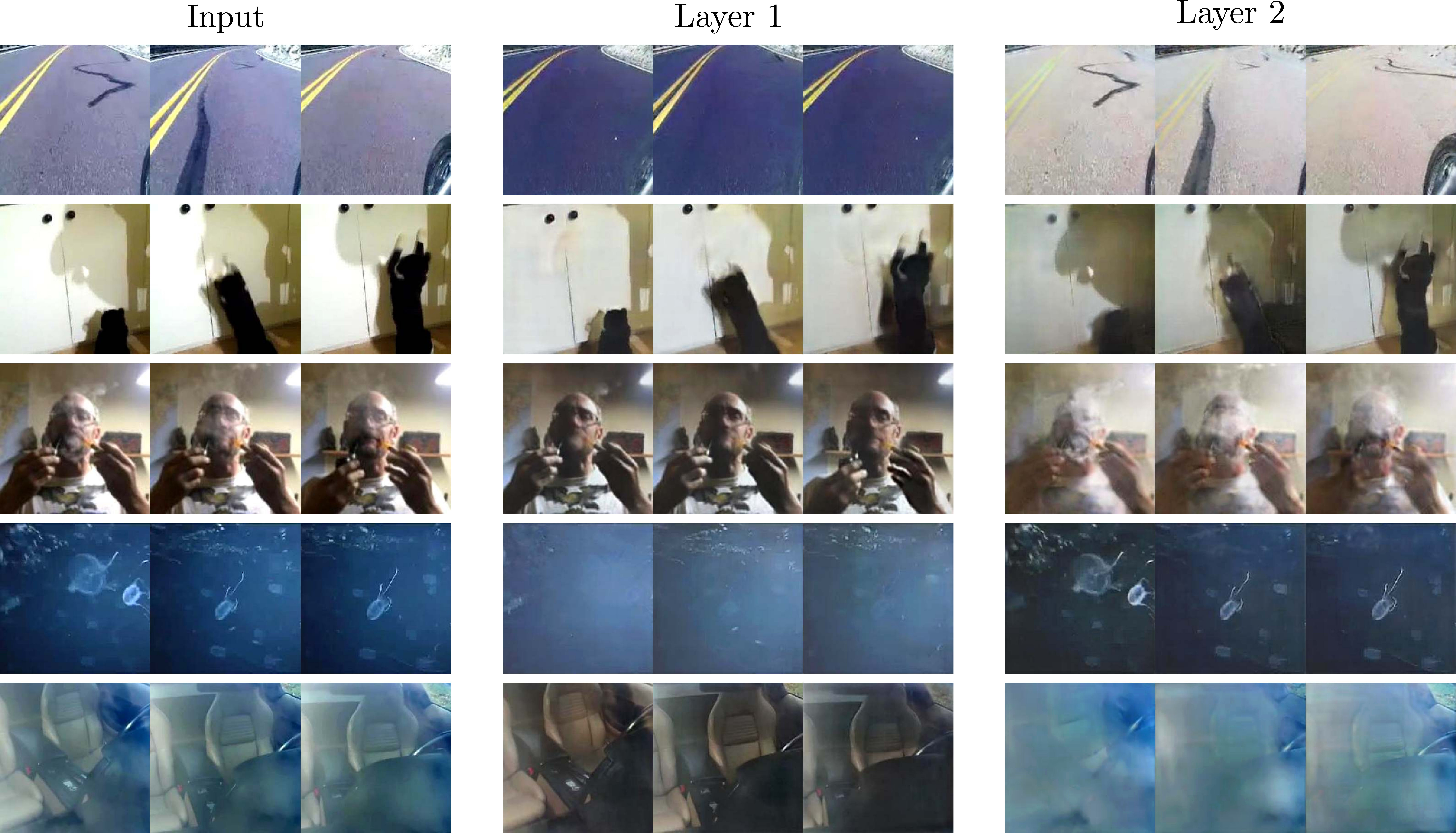}
	\end{center}
	\caption{\label{fig:qual_res_real} Results of our model on real-world videos containing transparency, reflections, shadows and even smoke.}
\end{figure*}

As a second experiment we ran the predictor-corrector model on blended videos formed of pairs from the same Kinetics human action classes, and found that the average loss was 0.145, higher than 0.133 when operating random pairs of videos. However this may also be explained by actions in the same class having similar low-level statistics.

As a third experiment we measured again the unmixing losses, but this time we recorded also two distances between each video in a pair that gets blended together, using euclidean distance on features from an I3D action classifier trained with supervision on Kinetics-600. One distance between low-level features (averaging features from the second convolutional layer) and the other between high-level features (averaging deeper "Mixed\_5c" features). We then measured the Pearson correlation between the losses and each of the two distances. We found a negative correlation of -0.23 between high-level distance and loss, confirming that videos showing similar (low-distance) actions tend to be hard to separate, but a weaker positive correlation between losses and low-level distances of 0.14, showing that low-level similarity is less of a challenge for unmixing.

\section{Applications}
\label{sec:applications}

In this section, we discuss the applicability of our method to real videos. For these experiments we trained the proposed model on 64-frame clips with 224x224 resolution. We first discuss the computational efficiency of our architecture in section~\ref{subsec:efficiency} before showing results on videos composed of various naturally layered phenomena such as reflections, shadows or occlusions in section~\ref{subsec:qual_res}.

\subsection{Efficiency}
\label{subsec:efficiency}
Our base network takes approximately 0.5 seconds to process a $64$-frame clip at $224\times 224$ resolution, using $4$ output layers. If we use our biggest model, the corrector-predictor, it then takes approximately twice that time. These timings are reported using a single P4000 Nvidia GPU. Note that this is significantly faster than the timings reported by techniques in related areas, such as for reflection removal ~\cite{Xue2015} which require minutes to process a similar video.
In addition, our network can seamlessly be applied to longer and higher definition videos as it is fully convolutional.

\subsection{Real world layer decomposition}
\label{subsec:qual_res}
\begin{figure}[t!]
	\begin{center}
		\includegraphics[width=\linewidth]{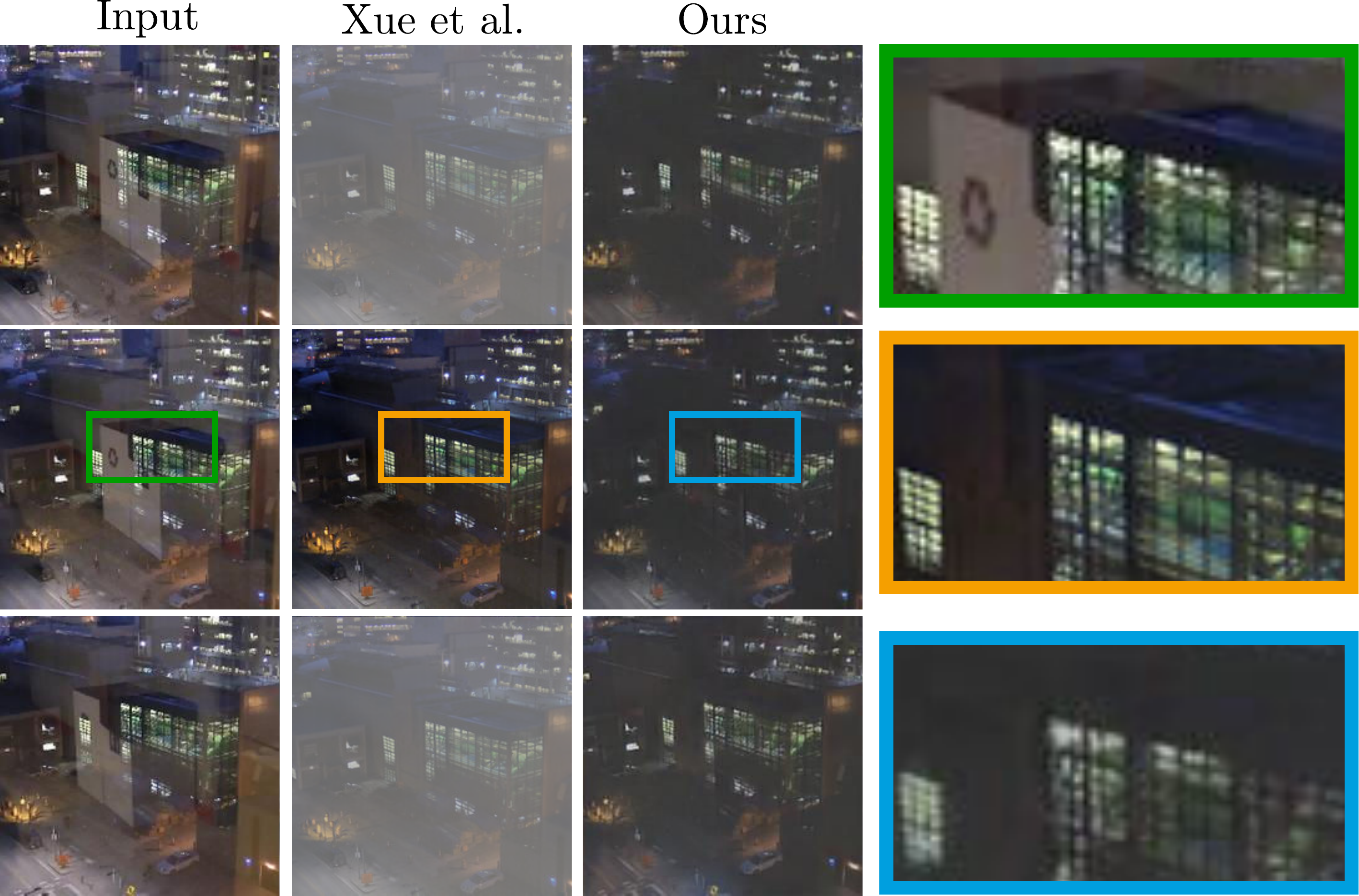}
	\end{center}
\vspace*{-0.5cm}
	\caption{\label{fig:qual_res_compar} Comparison of the centrifuge with a method specifically engineered for the purpose of reflection removal~\cite{Xue2015} (we unfortunately do not have their results for the first and third frames).}
	\vspace*{-0.5cm}
\end{figure}

We now demonstrate that, even if trained with synthetic videos, the proposed model is able to generalize to standard videos, sourced from the web.  A selection showcasing various types of natural video layers such as reflections, shadows and smoke is presented in \fig~\ref{fig:qual_res_real}. The model tends to perform quite well across many videos, in regions of the videos where such compositions do occur; outside those regions it sometimes distorts the videos (or perhaps we do not understand exactly what layers the model is considering). 
We also compare visually to a method that is specifically designed for reflection removal~\cite{Xue2015} in \fig~\ref{fig:qual_res_compar}.
Even if our results look less vivid than~\cite{Xue2015}, the centrifuge does a reasonable job at this task while making fewer assumptions.

\section{Conclusion}
\label{sec:conclusion}
We have presented a model that can be trained to reconstruct back individual videos that were synthetically mixed together, in the spirit of real-life centrifuges which separate materials into their different components. We explored what were the important bits to suceed at the training task, namely a permutation invariant loss, the depth of the network, the ability to produce multiple hypotheses and the recursive approach with our predictor-corrector model. We also investigated what are the cues used by our model and found evidence that it relies on both semantic \emph{and} low-level cues, especially motion. 

Our main scientific goal, however, was to find out what such a system would do when presented with a single (not synthetically mixed) video and we verified that it learns to tease apart shadows, reflections, and lighting. One can only hope that, as we look at the world through the lenses of more advanced models of this kind, we can uncover new layers of reality that are not immediately apparent, similar to what hardware-based advanced such as microscopes and telescopes have done in the past -- but now in the pattern recognition domain.

Much work remains to be done, in particular on how to control the
layer assignment process to make it more useful for applications,
which may include robust perceptual frontends for safety-critical
systems operating in complex visual scenes (e.g. self-driving cars) or
in video editing packages.  Future work should also consider relaxing the uniform mixing
of videos that we employed here -- both to make the learning problem harder but hopefully also
to improve the visual quality of the separated layers.

\noindent
\textbf{Acknowledgments.}	
We would like to thank Relja Arandjelovi\'c, Carl Doersch, Viorica Patraucean and Jacob Walker for valuable discussions, and the anonymous reviewers for very helpful comments.

{\small
\bibliographystyle{ieee}
\bibliography{shortstrings,egbib}
}

\clearpage
\appendix

\section{Details about model architectures} 
\label{app:architecture}
\begin{figure*}[t]
	\begin{center}
		\includegraphics[width=0.7\linewidth]{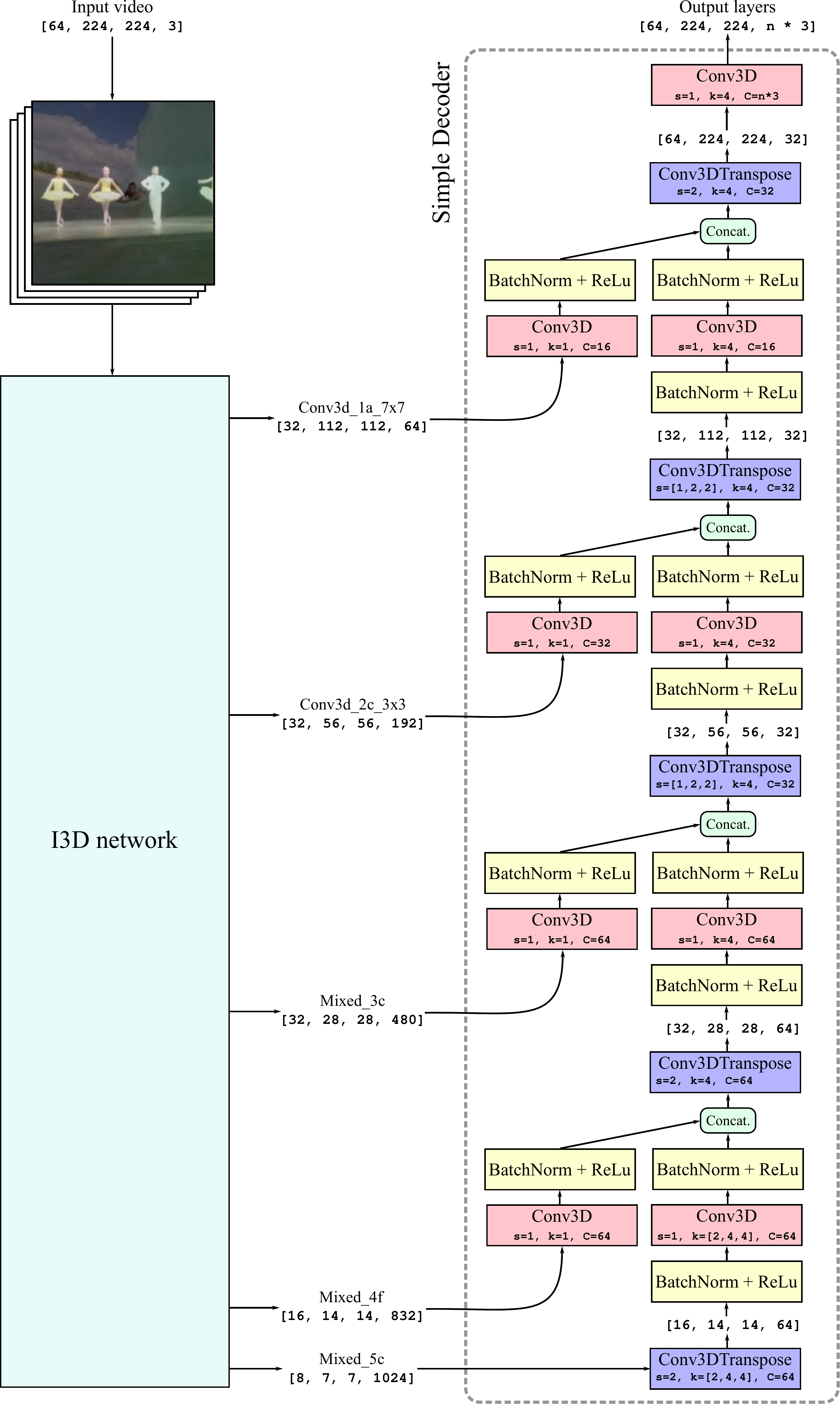}
	\end{center}
	\caption{\label{fig:architecture_predictor_decoder} \small 
		Details of our U-Net encoder-decoder architecture.	
		$\texttt{s}$ stands for stride, $\texttt{k}$ means kernel shape, $\texttt{C}$  gives number of channels.
		Finally, $n$ is the number of desired output layers.
	}
\end{figure*}

\noindent \textbf{Predictor.} Each \textit{predictor} module is composed of two submodules, an encoder and a decoder. For the encoder we used I3D~\cite{carreira17quovadis}, excluding its last pooling+classification layers. For the decoder we put together a simple 3D upconvolutional network, with skip connections coming from the encoder. The details of the decoder are provided in Fig.~\ref{fig:architecture_predictor_decoder}.

\begin{figure}[h]
	\begin{center}
		\includegraphics[width=0.5\linewidth]{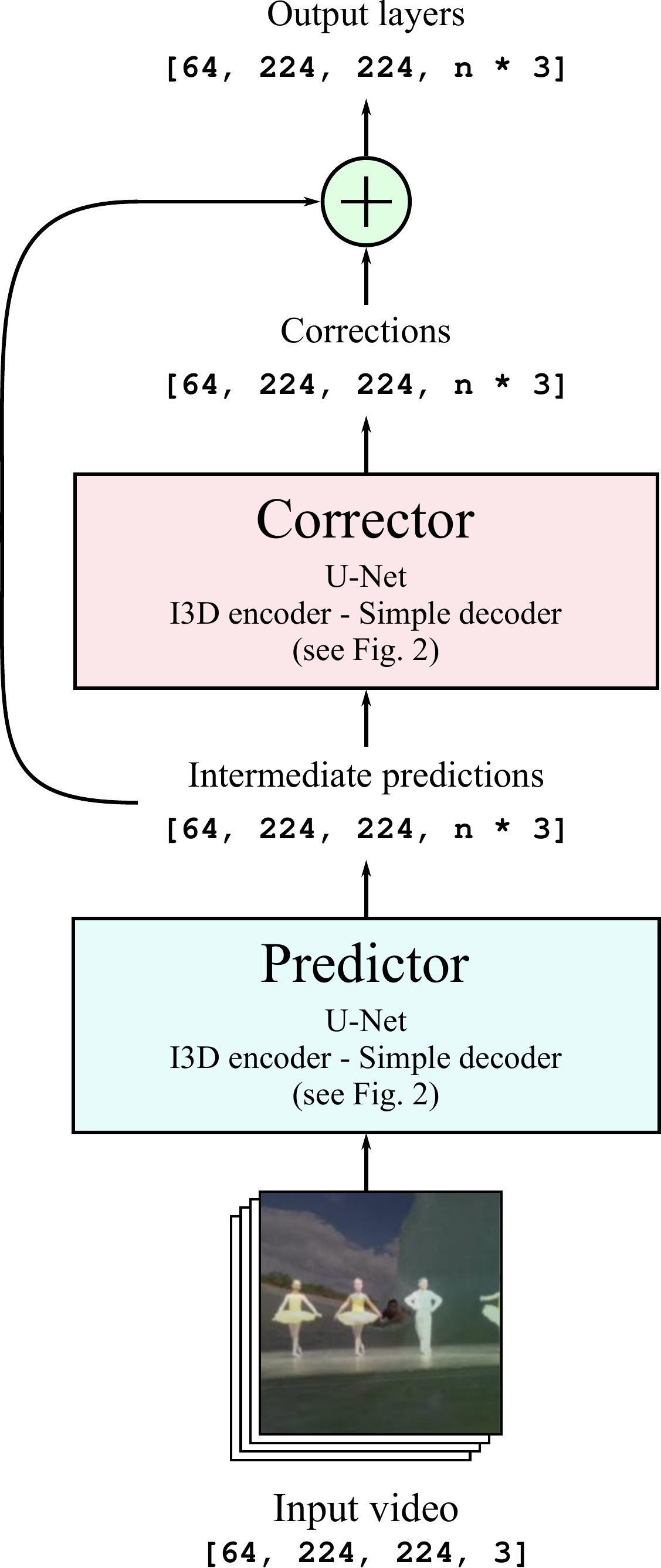}
	\end{center}
	\caption{\label{fig:architecture_predictor_corrector} \small 
		Predictor-Corrector architecture. $n$ specifies the number of desired output layers.
	}
\end{figure}

\vspace{3mm}
\noindent \textbf{Predictor-Corrector.} This model puts together two \textit{predictors}: the first one processes a single video. The second one processes the output of the first one, a stack of video layers (e.g. $4$ in most of our experiments, and produces corrections to each of them. The final video layers equal the sum of the outputs of both modules. This architecture is shown in Fig.~\ref{fig:architecture_predictor_corrector}.

\end{document}